%% file: main.tex
\documentclass[10pt,twocolumn,letterpaper]{article}

\usepackage[pagenumbers]{cvpr} 
\usepackage{multirow}
\usepackage{colortbl}
\usepackage{color}
\definecolor{LightCyan}{rgb}{0.88,1,1}
\input{preamble}

\usepackage{indentfirst}
\definecolor{cvprblue}{rgb}{0.21,0.49,0.74}
\usepackage[pagebackref,breaklinks,colorlinks,citecolor=cvprblue]{hyperref}
\def\paperID{} 
\def\confName{CVPR}
\def\confYear{2024}

\title{CLiF-VQA: Enhancing Video Quality Assessment by Incorporating High-Level Semantic Information related to Human Feelings}

\author{Yachun Mi, Yu Li, Yan Shu, Chen Hui, Puchao Zhou, Shaohui Liu\thanks{Corresponding author.}\\
Harbin Institute of Technology\\
}

\begin{document}
\maketitle
\input{sec/0_abstract}    
\input{sec/1_intro}
\input{sec/2_related}
\input{sec/3_clip}

\input{sec/4_approach}

\input{sec/5_experiments}
\input{sec/6_conclusions}
{
    \small
    \bibliographystyle{ieeenat_fullname}
    \bibliography{main}
}

\end{document}

%% file: preamble.tex
\usepackage[dvipsnames]{xcolor}


%% file: sec/0_abstract.tex
\begin{abstract}
Video Quality Assessment (VQA) aims to simulate the process of perceiving video quality by the human visual system (HVS).
The judgments made by HVS are always influenced by human subjective feelings.
However, most of the current VQA research focuses on capturing various distortions in the spatial and temporal domains of videos, while ignoring the impact of human feelings.
In this paper, we propose CLiF-VQA, which considers both features related to human feelings and spatial features of videos.
In order to effectively extract features related to human feelings from videos, we explore the consistency between CLIP and human feelings in video perception for the first time.
Specifically, we design multiple objective and subjective descriptions closely related to human feelings as prompts. Further we propose a novel CLIP-based semantic feature extractor (SFE) which extracts features related to human feelings by sliding over multiple regions of the video frame.
In addition, we further capture the low-level-aware features of the video through a spatial feature extraction module. The two different features are then aggregated thereby obtaining the quality score of the video.
Extensive experiments show that the proposed CLiF-VQA exhibits excellent performance on several VQA datasets.
\end{abstract}

%% file: sec/1_intro.tex
\section{Introduction}
\label{sec:1}
With the rapid advancement of technology, the threshold of video production has been significantly lowered, enabling an increasing number of users to create and upload videos to various online platforms (e.g., TikTok, YouTube).
However, user-generated content (UGC) videos often have annoying distortion because of the absence of professional filming equipment and skills.
Therefore, Video Quality Assessment (VQA) of in-the-wild videos is increasingly important for major video platforms to filter out and enhance low-quality videos.

\begin{figure}[t]
	\vspace{-0.4cm}
	\centering
	\includegraphics[width=8.3cm]{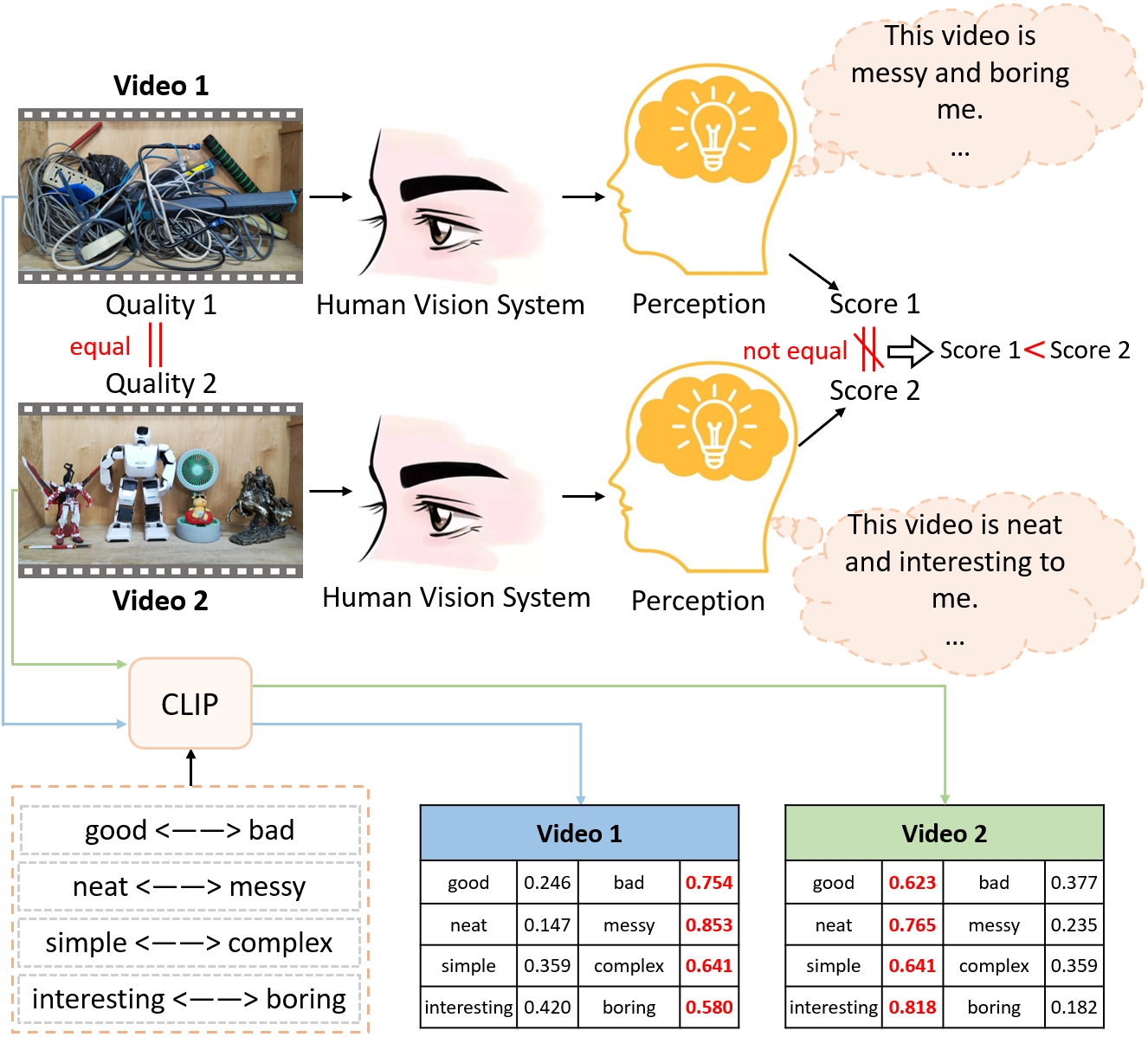}
	\caption{Validating the impact of human feelings on HVS for VQA and the relevance of CLIP with human feelings in video perception. Two videos with the same quality captured in the same scene using the same equipment.  
		CLIP show high consistency with human perception in video perception.
		We selected 10 subjects to perform VQA on the two videos and took their mean value as the video quality score. 
	}
	\label{fig1}
	\vspace{-0.4cm}
\end{figure}

The lack of raw information and the diversity of content and distortion types in in-the-wild videos present a significant challenge for VQA research.
Fortunately, there are many subjective experiments that provide high-quality datasets \cite{paper1,paper2,paper3,paper16,paper5,paper6}, which are labeled according to human mean opinion scores (MOS). 
With the benefit of these datasets, the current VQA methods can perform supervised training on them to fit the MOS as best as possible.
Traditional VQA methods \cite{paper7,paper8,paper9,paper10,paper11,paper12,paper13,paper14} are successful in predicting the quality of perceptual videos, which model spatial and temporal distortions using handcrafted features. 
However, the hand-crafted features have a low correlation with human perception, so its outcomes are not always reliable.
In recent years, with the advancement of deep learning techniques, VQA methods \cite{paper15,paper16,paper17,paper18,paper19} based on deep neural networks (DNNs) can extract more complex and abstract features related to video quality and achieve superior performance than traditional methods. 
However, most deep learning approaches focus on the effect of spatial and temporal video distortion on video quality, without adequately considering the relationship between video quality factors and human feelings.
It's pretty obvious that human judgments are always influenced by subjective feelings,  as shown in Fig. \ref{fig1}, two videos with the same quality but different content have different subjective quality scores.
Considering that all the current datasets used for VQA are labeled based on HVS, incorporating human feelings into VQA enables the model to achieve better consistency with HVS.
Although some recent studies \cite{paper21,paper15,paper22,paper23} have found that the content of the video affects the human judgment of the video quality, these studies model the effect of the content on subjectivity by extracting higher-level abstract features that are not directly related to human feelings, and thus these features do not effectively reflect human feelings.

However, extracting features related to human feelings from video is challenging due to the lack of research on how video affects human feelings.
Recently, Wang et al. \cite{paper25} revealed that Contrastive Language-Image Pre-training (CLIP) \cite{paper24} could evaluate the quality of images by using paired antonym prompts and achieved high consistency with human feelings on common IQA datasets \cite{paper26,paper27,paper28}.
This informs our study, but there are no studies showing that CLIP still has  perceptual abilities similar to human feelings in video.
Therefore, \textbf{an important contribution of this paper is to verify for the first time  that CLIP has a high degree of consistency with human feelings in video perception} through extensive experiments, as detailed in Sec. \ref{sec:3}. 
This opens up the possibility of incorporating human feelings into VQA tasks.


Based on the previous study, in this paper, we further propose a novel model to enhance video quality assessment (VQA) by incorporating high-level semantic information related to human feelings (denoted as CLiF-VQA), which further introduces high-level semantic features related to human feelings on top of the low-level-aware features extracted by FAST-VQA \cite{paper29}.
Specifically, we use a set of objective (e.g., bright, blurry, noisy, colorful, etc.) and subjective (e.g., pleasant, boring, fearful, etc.) descriptions that are closely related to human feelings as prompts. Further, we design a semantic feature extractor (SFE) which extracts high-level semantic features corresponding to the descriptions in multiple regions of the video frame.
Finally, we fuse the low-level-aware and high-level semantic features to obtain the video quality score.


Our contributions can be summarized as follows:

\begin{itemize}[leftmargin=2em,topsep=1pt,itemsep=6pt]
\item \textbf{We validate for the first time that CLIP is highly consistent with human feelings in video quality perception.} (Sec. \ref{sec:3})
\item \textbf{We propose CLiF-VQA, which incorporates for the first time features related to human feelings.}
It predicts the quality of a video by fusing high-level semantic features related to human feelings extracted by SEF and low-level perceptual features extracted by FAST-VQA. 
Extensive experiments demonstrate  that CLiF-VQA achieves the best performance on multiple VQA datasets.
\item \textbf{We design some efficient objective and subjective descriptions that are related to human feelings}. These descriptions are used as prompts for extracting high-level semantic information from the video.
\item \textbf{We design a zero-shot advanced semantic feature extractor (SFE) based on CLIP.} It extracts semantic features by sliding over multiple regions of a video frame and splices the same semantic features according to their relative positions to obtain semantic feature maps of the video frame. (Sec. \ref{sec:4})
\end{itemize}

%% file: sec/2_related.tex
\section{Related Works}
\label{sec:2}

\subsection{Classical VQA Methods}
Classical VQA methods employ handcrafted features to capture specific types of distortions in the video for quality prediction.
Early VQA often apply image quality assessment (IQA) algorithms \cite{paper30,paper31,paper32,paper38,paper39,paper40} to obtain frame-level features, and then combine with temporal dimension information to obtain video quality scores.
For example, V-CORNIA \cite{paper33} extends the IQA algorithm CORNIA \cite{paper32} to VQA to obtain frame-level quality scores, and combines these scores through temporal pooling.
However, this method does not fully consider the connection between the spatio-temporal information of the video and how they affect the video quality \cite{paper34,paper35,paper11,paper36}.
Natural video statistics (NVS) can take into account both spatio and temporal information, thus it is applied to address the previous problem.
V-BLIINDS \cite{paper11} extracts spatio-temporal statistical features of frame-differences in the video DCT domain and predicts crude frame quality scores using NIQE \cite{paper31}, then trains a linear kernel support vector regression (SVR) \cite{paper37}. 
TLVQM \cite{paper8} considers two levels of features, first computing low complexity features for each frame to extract frame-level statistical features related to motion, and then computing high complexity features related to spatial distortion for representative frames.
VIDEVAL \cite{paper7} applies various handcrafted features to detect and measure the distortions and reduces the computational complexity by reducing the feature dimensions.


\subsection{Deep Learning-based VQA Methods}
Recently, deep learning-based VQA Methods \cite{paper41,paper15,paper42,paper16,paper17,paper18,paper29,paper23,paper43,paper19,paper45,paper46,paper20,paper68,paper69} have gradually achieved better performance than classical methods. 
Rather than relying on handcrafted features, deep VQA methods employ convolutional neural networks (CNN) \cite{paper47,paper48,paper49,paper50,paper51,paper52,paper55,paper56} or Transformer models \cite{paper44,paper53,paper54} to extract complex and abstract features that are relevant to video quality aspects.
For example, VSFA \cite{paper15} extracts spatial features of video frames using ResNet-50 \cite{paper48} pre-trained on ImageNet \cite{paper57}, and then models the temporal features using GRU \cite{paper58}. 
Similar to the architecture of VSFA, while GST-VQA \cite{paper41} applies VGG-16 \cite{paper47} to extract spatial features of videos.
To better capture the spatio-temporal information of the video, some works \cite{paper16,paper18,paper22,paper43,paper62,paper42} adopt 3D-CNN. 
For example, V-MEON \cite{paper42} adopts a multi-task framework which utilizes 3D-CNN to extract spatio-temporal features to predict the quality of the video.
Other studies \cite{paper16,paper22,paper43,paper18} combine both 2D-CNN and 3D-CNN to capture the spatial and temporal features of video, and then integrate the two features for quality prediction.
Recently, VQA methods \cite{paper23,paper29,paper19} using the transformer structure have achieved better results relative to CNN. 
DisCoVQA \cite{paper23} uses Video Swin Transformer \cite{paper53} to extract multi-level spatio-temporal features and improves the learning efficiency of the model by temporal sampling of the features.
Similarly, FAST-VQA \cite{paper29} and FasterVQA \cite{paper19} obtain fragments by spatial-temporal grid mini-cube sampling (St-GMS) and then feed the fragments into a modified Video Swin Transformer \cite{paper53}.

Although deep learning-based VQA methods can extract complex high-level semantic features, these features are not directly related to the human point of view.
Two recent works \cite{paper63,paper64} attempt to address this issue.
Specifically, MaxVQA \cite{paper63} captures a variety of quality factors that can be observed by humans through a modified vision-language foundation model CLIP and can jointly evaluate multiple specific quality factors and overall perceptual quality scores.
Dover \cite{paper64} assesses video quality from both aesthetic and technical perspectives by collecting a large number of subjective human quality opinions, so it relatively well models the human process of perceiving quality.


%% file: sec/3_clip.tex
\section{CLIP for Video Perception}
\label{sec:3}

\begin{figure}[htb]
	
	\begin{minipage}[b]{1\linewidth}
		\centering
		\centerline{\includegraphics[width=7.5cm]{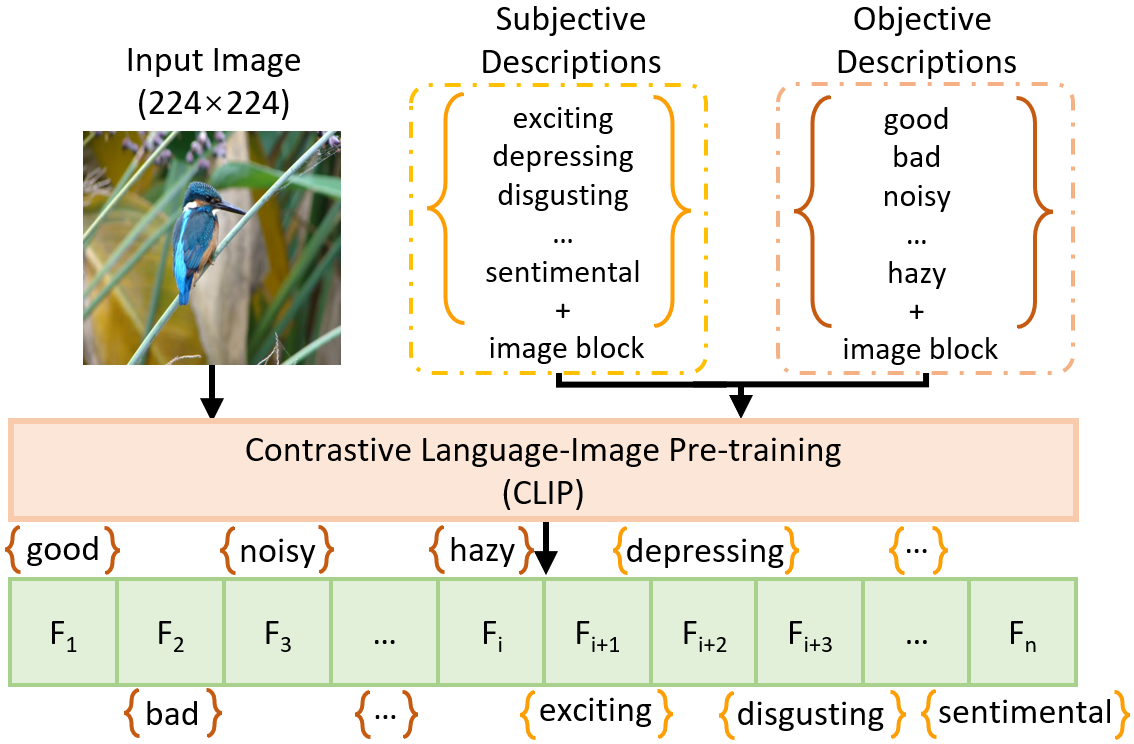}}
	\end{minipage}
	\caption{The process of extracting semantic features using CLIP.}
	\label{fig4}
\end{figure}

Recently, Wang et al. \cite{paper25} revealed that CLIP could evaluate the quality of images by using paired antonym prompts and achieved high consistency with human feelings on common IQA datasets \cite{paper26,paper27,paper28}.
Inspired by their work, we further explore the performance of CLIP in video perception. 
Unlike the pairwise antonym prompts strategy they used, we apply multiple objective descriptions related to quality factors (e.g., bright, contrast, etc.)  and multiple  subjective descriptions related to human feelings (e.g., interesting, exciting, etc.)  as prompts to obtain features in multiple dimensions of the video corresponding to the descriptions, as shown in Fig. \ref{fig4}.
We extract semantic features from multiple regions $ 224\times 224 $ with the help of CLIP on all video frames by means of sliding window. Then we compute the mean of all the feature values corresponding to a specific description as the feature of the video for that description.

Here, we refer to HVS's perception of video quality factors and content as objective and subjective feelings, respectively.
Further, we explore the correlation between CLIP and human objective feelings and subjective feelings.

\begin{figure}[htb]
	\begin{minipage}[b]{1\linewidth}
		\centering
		\centerline{\includegraphics[width=8.1cm]{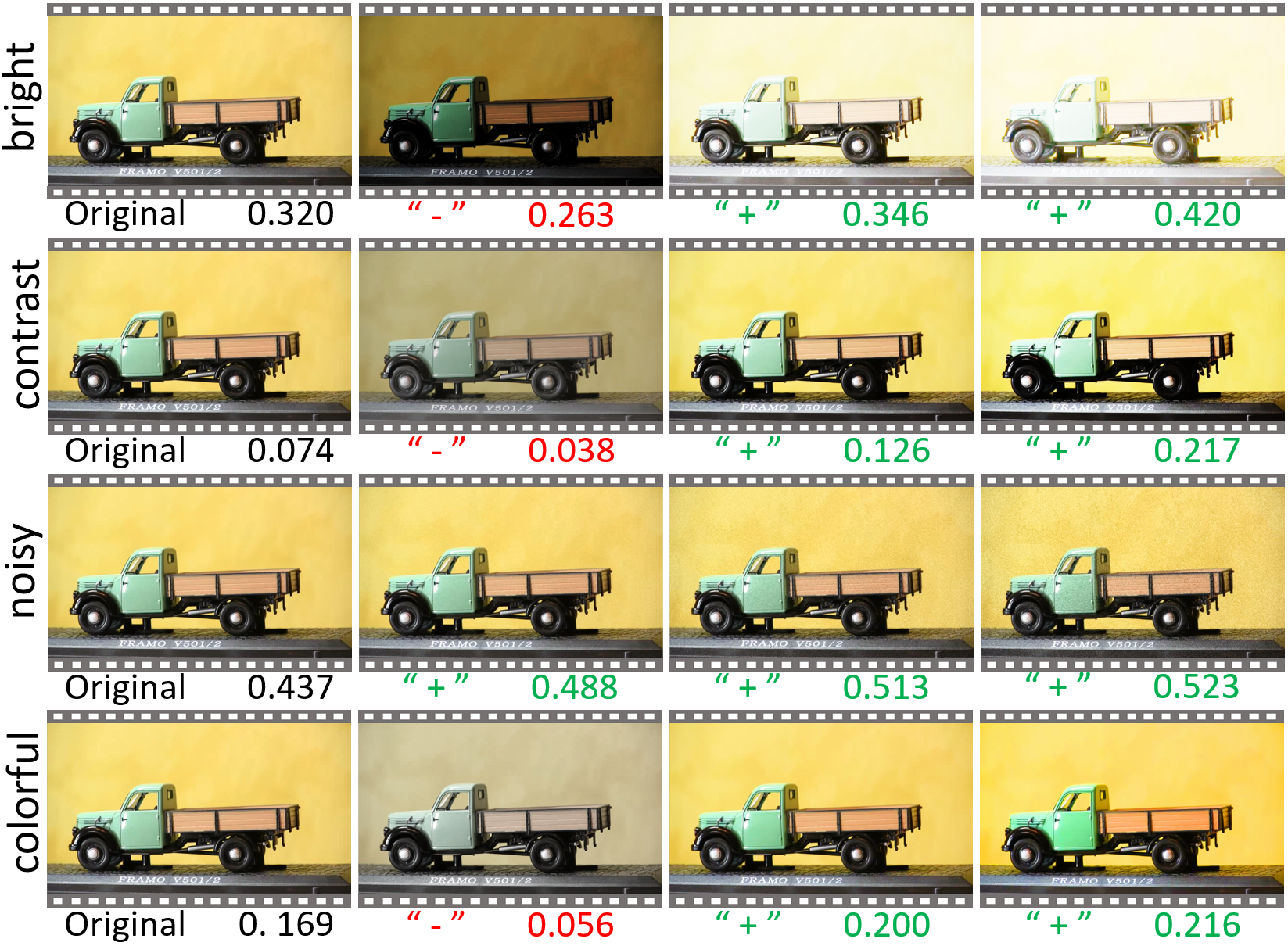}}
	\end{minipage}
	\caption{CLIP for perception of four objective descriptions (bright, contrast, noisy, colorful) related to video quality factors. "-" represents attenuation and "+" represents enhancement.}
	\label{fig2}
\end{figure}

\noindent
\textbf{Perception of Objective Feelings.}  We explore the performance of CLIP on four objective descriptions (bright, contrast, noisy, colorful) related to video quality factors, as shown in Fig. \ref{fig2}.
Specifically, we first process the video corresponding to a certain description, and then extract the semantic features  that correspond to the description.
It can be seen that CLIP is able to accurately perceive changes in video quality factors. 
This shows that CLIP has a good consistency with human objective feelings in the perception of video quality factors.


\begin{figure}[htb]
	\begin{minipage}[b]{1\linewidth}
		\centering
		\centerline{\includegraphics[width=8.1cm]{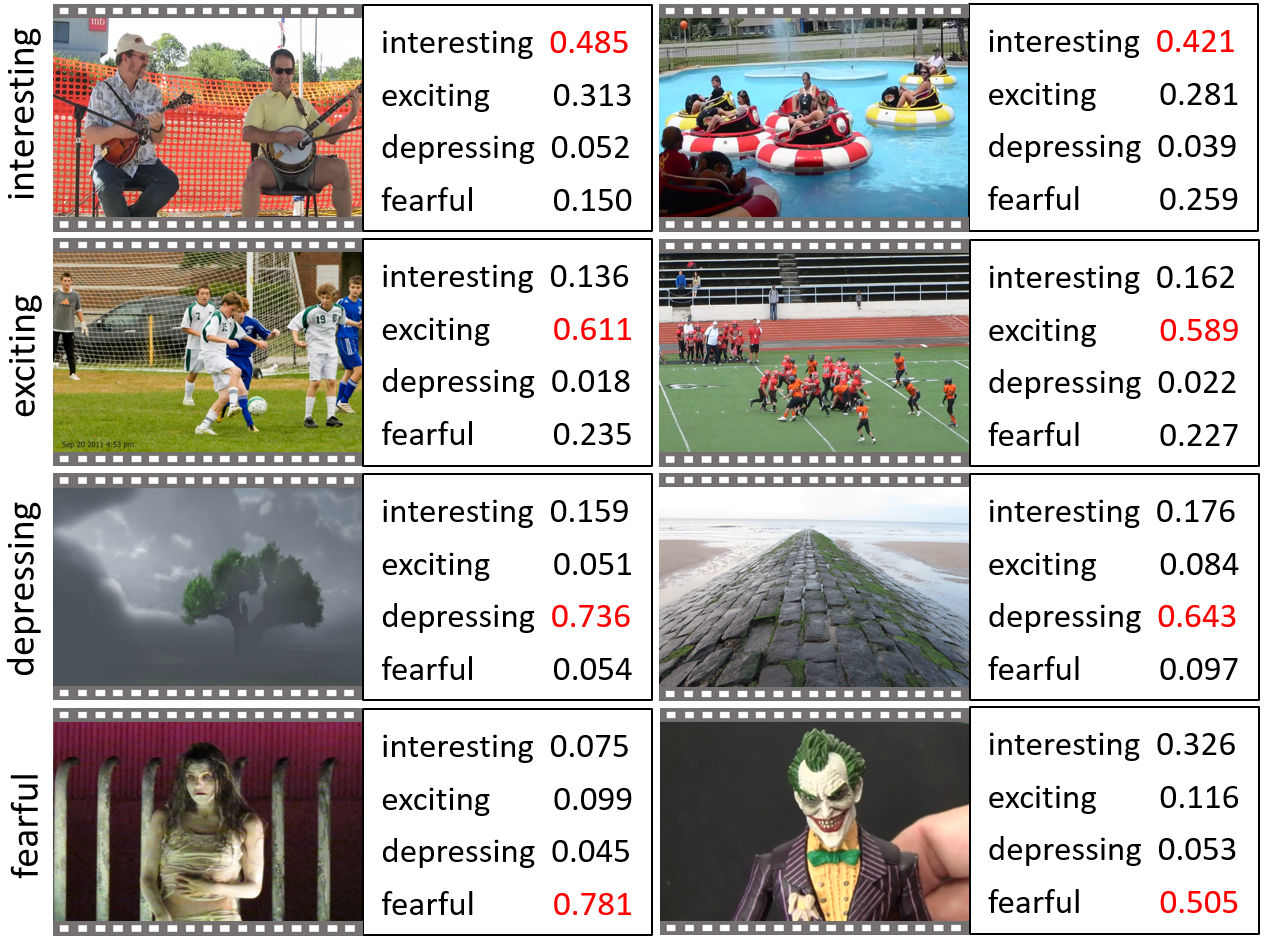}}
	\end{minipage}
	\caption{CLIP for perception of four subjective descriptions (interesting, exciting, depressing, fearful) related to human feelings.}
	\label{fig3}
\end{figure}

\noindent
\textbf{Perception of Subjective Feelings.}
Furthermore, we explore the relationship between CLIP and human subjective feelings in video content perception.
In particular, we conduct experiments on four subjective descriptions (interesting, exciting, depressing, fearful) that reflect the subjective feelings that video content brings to humans. As shown in Fig. \ref{fig3}. 
The results demonstrate that CLIP is highly consistent with human judgments in perceiving video content. 


\noindent
\textbf{Performance of CLIP in VQA.}
The experiments above demonstrate that CLIP has highly consistent results with humans in perceiving both the quality and content of the video separately.
However, it remains to be verified whether CLIP is still effective when both objective and subjective descriptions are used as prompts. 
Therefore, we conduct further experiments to explore CLIP's performance in video quality perception when using both objective and subjective descriptions that can reflect human feelings.

\begin{figure}[htb]
	
	\begin{minipage}[b]{1\linewidth}
		\centering
		\centerline{\includegraphics[width=8.1cm]{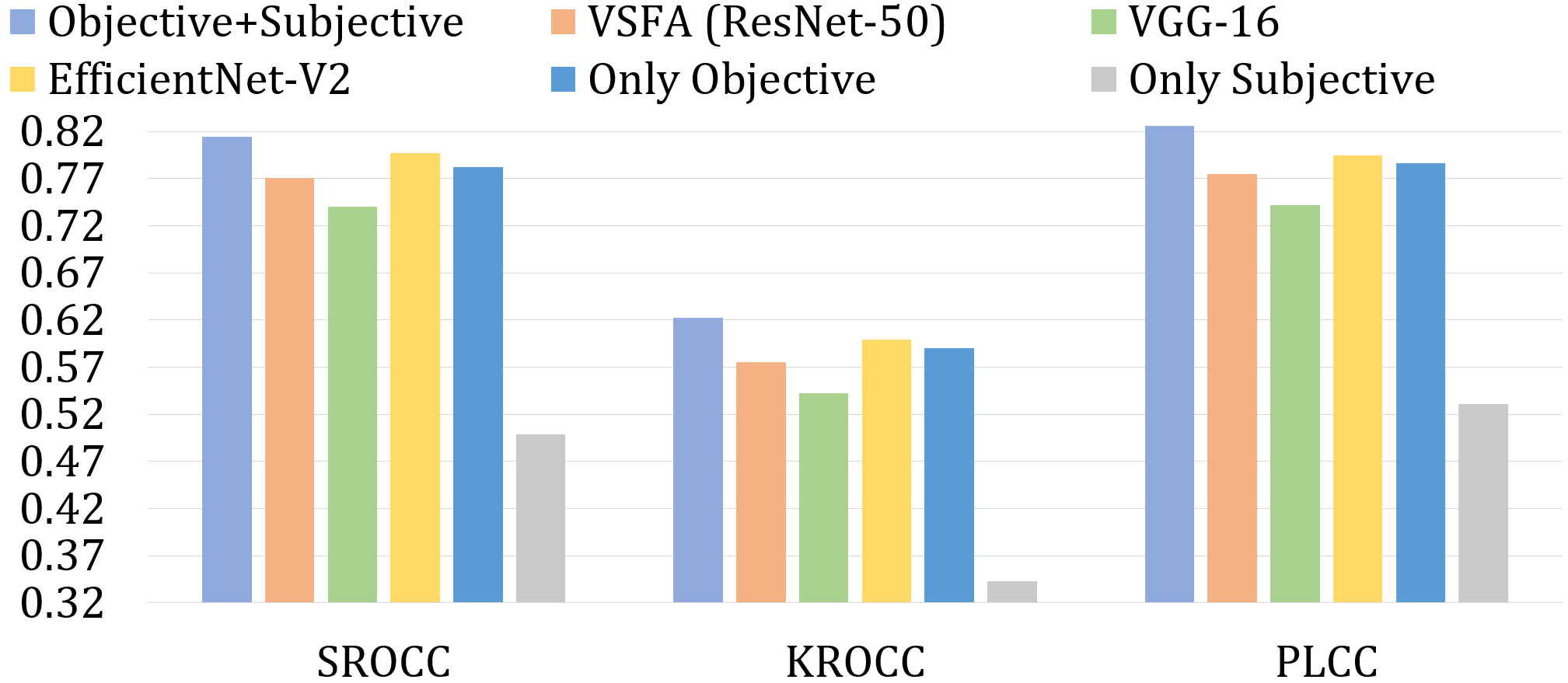}}
	\end{minipage}
	\caption{Performance using different descriptions as prompts and comparison results with some classical methods (ResNet-50 \cite{paper48}, VGG-16 \cite{paper47}, EfficientNet-V2 \cite{paper49}) on KoNViD-1k dataset .}
	\label{fig5}
	\vspace{-0.2cm}
\end{figure}

Specifically, we use multiple objective and subjective descriptions as prompts for CLIP, as shown in Fig. \ref{fig4} and then we still use the sliding window approach to extract semantic features on video frames, and finally we concatenate all the semantic features of each frame to obtain the feature vector of each frame of the video. Then we adopt the architecture of the classical VQA model VSFA \cite{paper15} to input the feature vectors of the frames into the GRU for regression and time pooling operation to get the quality scores. We conduct our experiments on the KoNViD-1k dataset \cite{paper1}, and in addition to processing the comparison with the VSFA model, we also compare with two feature extraction methods (VGG-16 \cite{paper47} and EfficientNet-V2 \cite{paper49}) widely used in VQA.
And we evaluate performance on SROCC,  KROCC and PLCC metrics, as shown in Fig. \ref{fig5}.
 
The results demonstrate that using both objective and subjective descriptions can achieve better results, compared to using a single description. 
Furthermore, it can be observed that relying on only features related to human feelings surpasses CNN extracted features in VQA. 
The experimental results also confirm the prompts of our design.

%% file: sec/4_approach.tex
\section{Approach}
\label{sec:4}

\begin{figure*}
	\vspace{-0.2cm}
	\centering
	\begin{subfigure}{0.68\linewidth}
		\centering
		\includegraphics[scale=0.19]{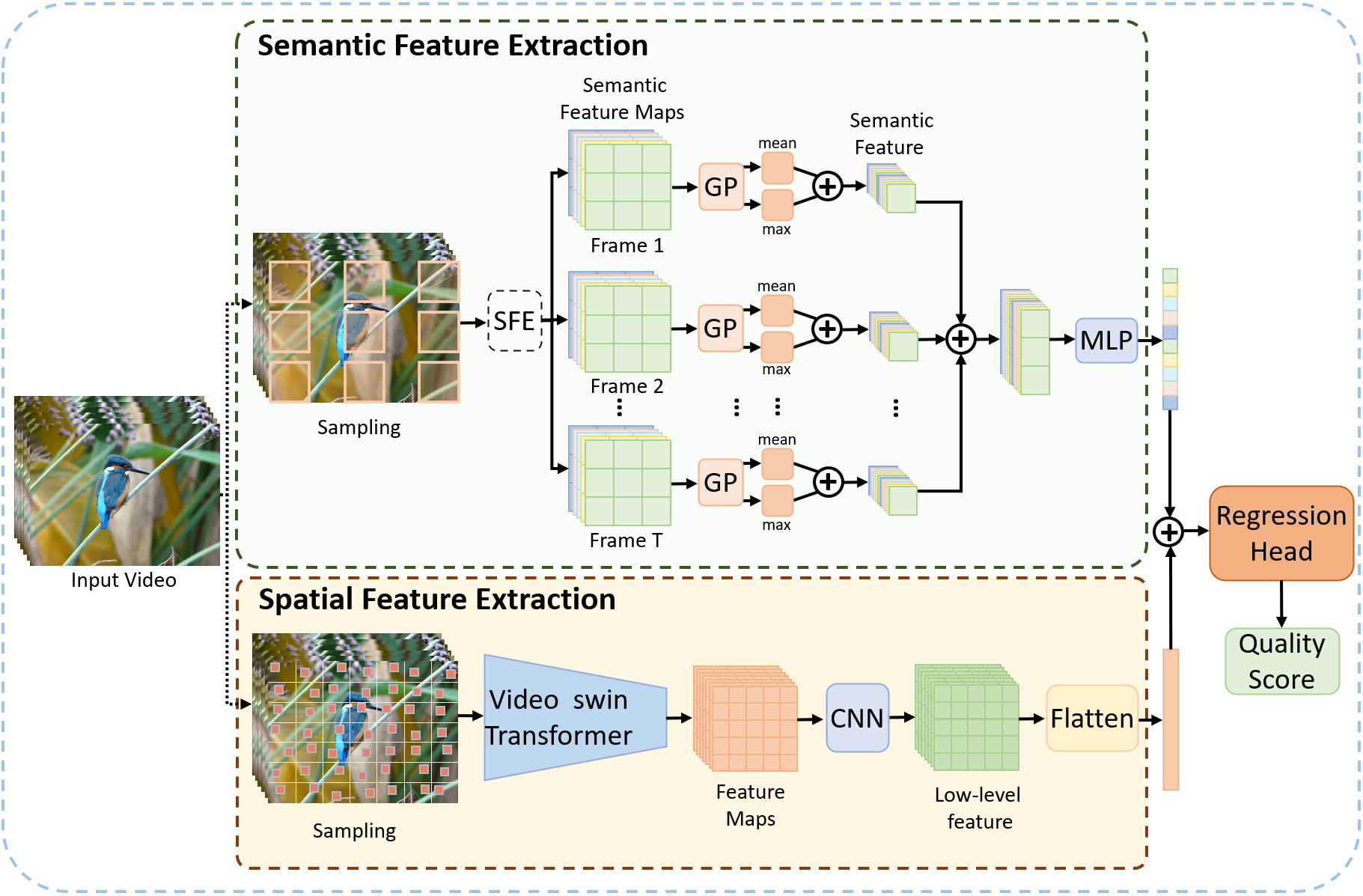}
		\caption{Illustration of the framework of CLiF-VQA.}
		\label{fig6-a}
	\end{subfigure}
	\hfill
	\begin{subfigure}{0.29\linewidth}
		\centering
		\includegraphics[scale=0.19]{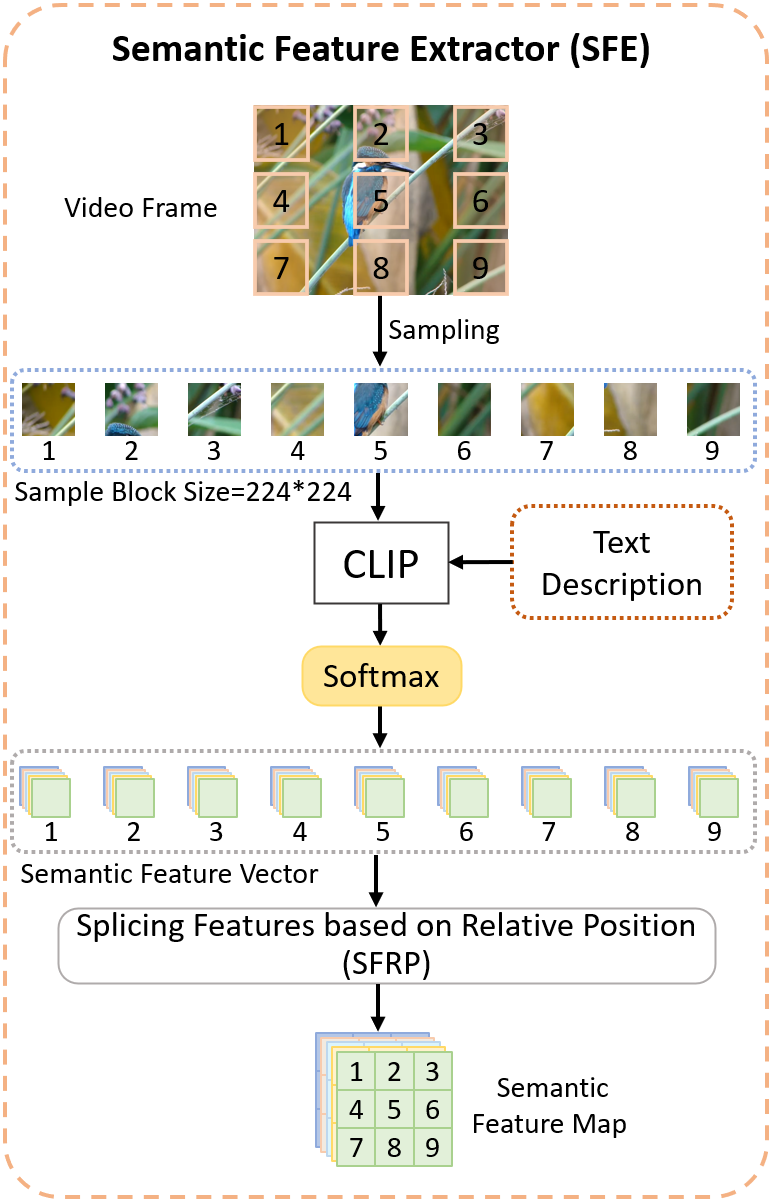}
		\caption{Semantic feature extractor (SFE). }
		\label{fig6-b}
	\end{subfigure}
	\caption{The framework of our proposed method, which extracts semantic features related to human feelings through the semantic feature extraction module as well as low-level perceptual quality features through a spatial feature extraction module, and then obtains the quality scores by aggregating the two features through the aggregation header.}
	\label{fig6}
	\vspace{-0.2cm}
\end{figure*}

In this section, we introduce the proposed CLiF-VQA, which consists of the semantic feature extraction module \ref{sec:4.1} and the spatial feature extraction module \ref{sec:4.2}, as shown in Fig. \ref{fig6}.
First, we employ the semantic feature extraction module to extract high-level semantic features that are related to human feelings. 
Then low-level-aware features are extracted using spatial feature extraction module.
Finally we fuse these two features through a regression module thus obtaining the video quality score.

\subsection{Semantic Feature Extraction}
\label{sec:4.1}
In order to effectively extract features that can reflect human feelings, we first design some objective descriptions and subjective descriptions related to human feelings as prompts of CLIP, as shown in Fig. \ref{fig4}. 
In addition, due to the limitation of the visual coder of CLIP on the input size, we can only extract the semantic information of a small region in the video frame.
In order to be able to obtain as much semantic information as possible contained in the video frames, we extracted features from multiple regions of the video frames by sampling them multiple times at different locations, as shown in Fig. \ref{fig6-b}. 
This avoids the loss of video quality caused by resizing and excessive cropping.

Specifically, assuming the video has $T$ frames, we perform a sampling operation on the video frames $ I_{t}(t = 1, 2, ..., T)$to obtain $m\times n$ image blocks of size $224\times 224$:
\begin{equation}
	\left \{ b_{t}^{i,j}|1 \le i\le m, 1 \le j\le n \right \} = Sampling(I_{t} )
\end{equation}
where $b_{t}^{i,j}$ represents the block obtained by sampling in the i-th row and j-th column.

CLIP calculates the cosine similarity between visual content and descriptions to predict scores for each dimension.
We use CLIP to extract high-level semantic features of $b_{t}^{i,j}$ related to human feelings.
:
\begin{equation}
	v_{t}^{i,j}=Softmax(CLIP(b_{t}^{i,j}),descriptions) 
\end{equation}
where descriptions are $r$ prompts of CLIP, and the feature values of $v_{t}^{i,j}$ are the same number and one-to-one correspondence with the number of descriptions.
$Softmax(\cdot)$ is to normalize the output of CLIP.

Then the splicing operation (SFRP) is performed on all the features $b_{t}^{i,j} $ based on the relative position to obtain the semantic feature map $M_{t}$  of frame $I_{t}$:
\begin{equation}
	M_{t} =SFRP(\left \{ v_{t}^{i,j}|1 \le i\le m, 1 \le j\le n \right \}) 
\end{equation}

$ M_{t} $ contains $r$ feature maps, each corresponding to a description. 
Further, we perform a global average pooling operation $(GP_{avg})$ and a global max pooling operation $(GP_{max})$ on $M_{t}$ to obtain the universal features and distinctive features as shown Fig. \ref{fig6-a}. 
The outputs of $(GP_{avg})$ and $(GP_{max})$ are two r-dimensional feature vectors $f_{t}^{avg}$  and $f_{t}^{max}$, respectively.
\begin{equation}
	f_{t}^{avg} = GP_{avg} (M_{t} )
\end{equation}
\begin{equation}
	f_{t}^{max} = GP_{max} (M_{t} )
\end{equation}

$f_{t}^{avg}$ and $	f_{t}^{max}$ are then concatenated as the semantic feature vectors $f_{t}$  of the video frame $I_{t}$:
\begin{equation}
	\label{ft}
	f_{t}=f_{t}^{avg}\oplus f_{t}^{max}
\end{equation}
where $\oplus$ is the concatenation operator and $f_{t}$ is a feature vector of length $2\times r$.

Next, we perform a concatenation operation on the semantic features $\left \{  f_{t}\right \}_{t=1}^{T} $ of all the video frames thereby obtaining the semantic feature maps $M_{s}$ of the video:
\begin{equation}
M_{s}= f_{1} \oplus f_{2} \oplus f_{3}\oplus ...\oplus f_{T}
\end{equation}
where $ \oplus $ here is not exactly the same as the concatenation of $ \oplus $ in Eq. \ref{ft}, here $ \oplus $ performs a parallel concatenation on the feature vectors $\left \{  f_{t}\right \}_{t=1}^{T} $ so that the dimension of the obtained $M_{s}$ is $[2\times r,T]$.
Each feature map corresponds to a description, and the feature maps here are divided into two types, namely the feature map with global average pooling operation $(GP_{avg})$ and the feature map with global max pooling operation $(GP_{max})$. 

After extracting the semantic feature maps of the video, we use a multi-layer perceptron (MLP) to obtain the feature vectors $F_{s}$ corresponding to the descriptions.The MLP is composed of two fully connected layers and the activation function is GELU:
\begin{equation}
F_{s}=FC_{2}(GELU(FC_{1}(M_{s} ) ))
\end{equation}
where $ FC_{1} $ and $ FC_{2} $ are two fully connected layers and $ F_{s} $ is a $ 2\times r $ dimensional feature vector.

\subsection{Spatial Feature Extraction}
\label{sec:4.2}
In VQA, the spatial features of the video play a very important role in estimating the overall video quality. Since low-level information is easily affected by distortion, extracting the low-level-aware features of the video can effectively capture the spatial distortion of the video.

In our approach, we apply FAST-VQA \cite{paper29} to perceive the low-level quality information in the video. 
FAST-VQA is a state-of-the-art VQA architecture, which has been proved to be effective in perceiving video distortions and has demonstrated outstanding performance on multiple VQA datasets.
Specifically, a new fragments sampling strategy is applied to obtain video fragments $ V_{f} $, which preserves the distortion information of the video $ V $ by splicing multiple small patches cropped at the original resolution.
Since continuous frame sampling is applied during the sampling process, $ V_{f} $ preserves the temporal distortion to some extent.
The video fragments $ V_{f} $ are then fed into a modified Video Swin Transformer Tiny \cite{paper53}:
\begin{equation}
	M_{f} = Transformer(V_{f})
\end{equation}
Next, the obtained features $ M_{f} $ are further processed by two 3D convolutional layers to obtain the final feature maps $ M_{final} $ of the video.
\begin{equation}
	M_{final}=Conv3D_{2}(GELU(Conv3D_{1}(M_{f})))
\end{equation}

Finally we flatten the feature maps $ M_{final} $ to obtain the spatial feature vector $ F_{f} $ of the video.
\begin{equation}
	F_{f}=Flatten(M_{final})
\end{equation}

In addition, although FAST-VQA can extract effective spatial quality perceptual features in the video, most of the video content is lost due to the segments sampling of FAST-VQA, thus seriously destroying the important information in the video related to human feelings. 
Therefore, the semantic features related to human feelings extracted in Sec. \ref{sec:4.1} can exactly complement the spatial features extracted by FAST-VQA.

\subsection{Quality Regression}
After extracting the semantic and spatial features of the video through the semantic feature extraction module and spatial feature extraction module, we need to map these features to the quality scores via a regression model.
First, We concatenate the semantic features $ F_{s} $ and spatial features $ F_{f} $ to get the overall features $ F_{v} $ of the video:
\begin{equation}
	F_{v} = F_{s} \oplus F_{f}
\end{equation}
Then we design a regression head with two fully connected layers to predict the quality score of the video:
\begin{equation}
	Score=FC_{4}(GELU(FC_{3}(F_{v})))
\end{equation}

\subsection{Loss Function}
The loss function used to optimize the proposed models consists of two parts: the monotonicity-induced loss and linearity-induced loss.
Given m predicted quality scores $ \hat{Q}=\left \{ \hat{q_{1}},\hat{q_{2}},...,\hat{q_{m}} \right \} $ and m ground-truth subjective quality scores $ Q=\left \{ q_{1},q_{2},...,q_{m} \right \} $.

Specifically, the monotonicity-induced loss predicts the monotonicity of the video quality scores by introducing additional order constraints. 
The monotonicity-induced loss function is defined as follows:
\begin{equation}
	L_{mon} =\frac{1}{m^{2} } \sum_{i=1}^{m} \sum_{j=1}^{m}max(0,\left | q_{i}- q_{j} \right |-f(q_{i}, q_{j})\cdot (\hat{q_{i}}-\hat{q_{j}} ) ) 
\end{equation}
where $ f(q_{i}, q_{j})=1 $ if $ q_{i}\ge q_{j} $, otherwise $ f(q_{i}, q_{j})=-1 $.

In contrast, the goal of the linearity-induced loss is to compute the linear relationship between the predicted quality score and ground-truth subjective quality score.
The linearity-induced loss function can be denoted as:
\begin{equation}
 ( 1-\frac{ {\textstyle \sum_{i=1}^{m}(\hat{q}_{i}-\hat{a} )({q}_{i}-{a})} }{\sqrt{ {\textstyle \sum_{i=1}^m{(\hat{q}_{i}- \hat{a})^{2} {\textstyle \sum_{i=1}^{m}(q_{i}-a)^{2}}  }} } } )/2 
\end{equation}
where $ a= \frac{1}{m} {\textstyle \sum_{i=1}^{m}} q_{i} $ and $ \hat{a} = \frac{1}{m} {\textstyle \sum_{i=1}^{m}} \hat{q}_{i} $.

Finally, the total loss function $ L $ is obtained by combining the two loss functions $ L_{mon} $ and $ L_{lin} $ above:
\begin{equation}
	L=\alpha L_{mon}+\beta L_{lin}
\end{equation}
where $ \alpha $ and $ \beta $ represent the weights of monotonicity-induced loss and linearity-induced loss.

%% file: sec/5_experiments.tex
\section{Experiments}
\label{sec:5}

\begin{table*}[t]
	\caption{Experimental performance of the pre-trained CLiF-VQA model on the LSVQ dataset on four test sets (LSVQ$_{test}$, LSVQ$_{1080p}$, KoNViD-1k, LIVE-VQC). 
		LSVQ$_{test}$ and LSVQ$_{1080p}$ are used for intra-dataset testing.
		While KoNViD-1k and LIVE-VQC are used for cross-dataset testing. 
		Best in \textbf{\textcolor{red}{red}} and second in \textbf{\textcolor{blue}{blue}}.}
	\centering
	\label{table:1}
	\small
	\begin{tabular}{l|c|cc|cc|cc|cc}
		\bottomrule
		\multicolumn{2}{c|}{Testing Type} & \multicolumn{4}{c|}{Intra-dataset Test Sets} & \multicolumn{4}{c}{Cross-dataset Test Sets} \\
		\hline
		\multicolumn{2}{c|}{Texting set} & \multicolumn{2}{c|}{LSVQ$_{test}$} & \multicolumn{2}{c|}{LSVQ$_{1080p}$} & \multicolumn{2}{c|}{KoNViD-1k} & \multicolumn{2}{c}{LIVE-VQC} \\
		\hline
		Methods & Source & SROCC & PLCC & SROCC & PLCC & SROCC & PLCC & SROCC & PLCC \\
		\hline
		BRISQUE \cite{paper30} & TIP, 2012 &  0.569 & 0.576  & 0.497  & 0.531  & 0.646  & 0.647  & 0.524  & 0.536 \\ 
		TLVQM \cite{paper8}& TIP, 2019 & 0.772 & 0.774  & 0.589  & 0.616  & 0.732  & 0.724  & 0.670  & 0.691 \\ 
		VIDEVAL \cite{paper7} & TIP, 2021 &  0.794 & 0.783  & 0.545  & 0.554  & 0.751  & 0.741  & 0.630  & 0.640 \\ 
		VSFA \cite{paper15}& ACMMM, 2019 & 0.801 & 0.796 & 0.675  & 0.704  & 0.784  & 0.794  & 0.734  & 0.772 \\ 
		PVQ$_{wo/patch}$ \cite{paper16} & CVPR, 2021 &  0.814 & 0.816  & 0.686  & 0.708  & 0.781  & 0.781  & 0.747  & 0.776 \\ 
		PVQ$_{w/patch}$ \cite{paper16} & CVPR, 2021 &  0.827 & 0.828  & 0.711  & 0.739  & 0.791  & 0.795  & 0.770  & 0.807 \\
		BVQA \cite{paper17} & TCSVT, 2022 &  0.852 & 0.854  & 0.771  & 0.782  & 0.834  & 0.837  & 0.816  & 0.824 \\
		FAST-VQA-M \cite{paper29}& ECCV, 2022 &  0.852 & 0.854  & 0.739  & 0.773  & 0.841  & 0.832  & 0.788  & 0.810 \\
		FasterVQA \cite{paper19}  & TPAMI, 2023 &  0.873 & 0.874  & 0.772  & 0.811  & 0.863  & 0.863  & 0.813  & 0.837 \\
		
		FAST-VQA \cite{paper29}& ECCV, 2022 &  0.872 & 0.874  & 0.770  & 0.809  & 0.864  & 0.862  & \textbf{\textcolor{blue}{0.824}}  & \textbf{\textcolor{blue}{0.841}} \\
		DOVER  \cite{paper64}& ICCV, 2023 &  \textbf{\textcolor{blue}{0.881}} & \textbf{\textcolor{blue}{0.879}}  & \textbf{\textcolor{blue}{0.782}}  & \textbf{\textcolor{blue}{0.827}}  & \textbf{\textcolor{red}{0.879}}  & \textbf{\textcolor{red}{0.885}}  & 0.807  & \textbf{\textcolor{blue}{0.841}} \\
		\hline
		\textbf{CLiF-VQA}  & \textbf{Ours} &  \textbf{\textcolor{red}{0.882}} & \textbf{\textcolor{red}{0.882}}  & \textbf{\textcolor{red}{0.788}}  & \textbf{\textcolor{red}{0.830}}  & \textbf{\textcolor{blue}{0.870}}  & \textbf{\textcolor{blue}{0.866}}  & \textbf{\textcolor{red}{0.835}}  & \textbf{\textcolor{red}{0.856}} \\
		\toprule
	\end{tabular}
\end{table*}

\subsection{Experimental Setups}

\noindent
\textbf{Datasets.} 
We test the proposed model on four datasets including LSVQ \cite{paper16}, KoNViD-1k (1200 videos) \cite{paper1}, LIVE-VQC (585 videos) \cite{paper3}, and YouTube-UGC \cite{paper6}. 
Specifically, we pre-train CLiF-VQA on LSVQ$_{train}$, a subset of  LSVQ containing 28,056 videos. 

Intra-dataset testing is performed on two subsets of LSVQ, LSVQ$_{test}$ (7400 videos) and LSVQ$_{1080p}$ (3600 videos). KoNViD-1k and LIVE-VQC 
And we perform cross-dataset testing on KoNViD-1k and LIVE-VQC.
Further, we fine-tune the model on  KoNViD-1k, LIVE-VQC, and YouTube-UGC.
It should be noted that  YouTube-UGC contains 1500 videos, but only 1147 videos are available to us.

\noindent
\textbf{Evaluation Criteria.} Spearman Rank Order Correlation Coefficient (SROCC) and Pearson Linear Correlation Coefficient (PLCC) are used as evaluation Metrics. 
Specifically, SRCC is used to measure the prediction monotonicity between predicted scores and true scores by ranking the values in both series and calculating the linear correlation between the two ranked series. 
In contrast, PLCC evaluates prediction accuracy by calculating the linear correlation between a series of predicted scores and true scores.
And higher SROCC and PLCC scores indicate better performance.

\noindent
\textbf{Implementation Details.} 
we employ PyTorch framework and an NVIDIA GeForce RTX 3090 card to train the model in all experimental implementations.
In the semantic feature extraction module, we sample each frame of the video 9 times and then perform zero-shot feature extraction with CLIP, as shown in Fig. \ref{fig6-b}.
In the spatial feature extraction module, we employ the FAST-VQA \cite{paper29} architecture with the Video Swin Transformer Tiny \cite{paper53} backbone pre-trained on the Kinetics-400 \cite{paper65} dataset.
During training, the initial learning rate of FAST-VQA  backbone is set to 0.000075, and the initial learning of other parts is set to 0.00075.
We set the batch size to 12 and use the AdamW optimizer with a weight decay rate of 0.05.

\begin{table*}[t]
	\caption{The finetune results on LIVE-VQC, KoNViD and YouTube-UGC datasets.
		CLiF-VQA* indicates that only the FAST-VQA branch is pre-trained on LSVQ, instead of training the whole model. CLiF-VQA* pre-trained only the spatial feature extraction branch on LSVQ.
		Best in \textbf{\textcolor{red}{red}} and second in \textbf{\textcolor{blue}{blue}}.
	}
	\centering\setlength{\tabcolsep}{10.5pt}
	\label{table:2}
	\small
	\begin{tabular}{l|c|cc|cc|cc}
		\bottomrule
		\multicolumn{2}{c|}{Finetune Datasets} & \multicolumn{2}{c|}{LIVE-VQC} & \multicolumn{2}{c|}{KoNViD-1k} & \multicolumn{2}{c}{YouTube-UGC}  \\
		\hline
		Methods & Source & SROCC & PLCC & SROCC & PLCC & SROCC & PLCC  \\
		\hline
		TLVQM \cite{paper8}& TIP, 2019 & 0.799 & 0.803  & 0.773  & 0.768  & 0.669  & 0.659  \\ 
		VIDEVAL \cite{paper7} & TIP, 2021 &  0.752 & 0.751  & 0.783  & 0.780  & 0.779  & 0.773  \\ 
		RAPIQUE \cite{paper66} & OJSP, 2021 &  0.755 & 0.786  & 0.803  & 0.817  & 0.759  & 0.768  \\ 
		CNN+TLVQM \cite{paper67} & ACMMM, 2020 & 0.825 & 0.834 & 0.816 & 0.818 & 0.809 & 0.802 \\
		CNN+VIDEVAL \cite{paper7} & TIP, 2021 & 0.785 & 0.810 & 0.815 & 0.817 & 0.808 & 0.803 \\
		VSFA \cite{paper15}& ACMMM, 2019 & 0.773 & 0.795 & 0.773  & 0.775  & 0.724  & 0.743  \\ 
		PVQ \cite{paper16} & CVPR, 2021 &  0.827 & 0.837  & 0.791  & 0.786  & NA  & NA  \\ 
		GST-VQA \cite{paper41} & TCSVT, 2021 & NA & NA & 0.814 & 0.825 & NA & NA \\ 
		CoINVQ \cite{paper22} & TCSVT, 2021 & NA & NA & 0.767 & 0.764 & 0.816 & 0.802\\ 
		BVQA \cite{paper17} & TCSVT, 2022 &  0.831 & 0.842  & 0.834  & 0.836  & 0.831  & 0.819 \\
		FAST-VQA-M \cite{paper29}& ECCV, 2022 &  0.803 & 0.828  & 0.873  & 0.872  & 0.768  & 0.765 \\
		FasterVQA \cite{paper19}  & TPAMI, 2023 &  0.843 & 0.858  & 0.895  & 0.898  & 0.863  & 0.859  \\
		FAST-VQA \cite{paper29} & ECCV, 2022 &  0.845 & 0.852  & 0.890  & 0.889  &  0.857  & 0.853  \\
		DOVER \cite{paper64} & ICCV, 2023 &  0.812 & 0.852  & \textbf{\textcolor{blue}{0.897}}  & \textbf{\textcolor{blue}{0.899}}  & \textbf{\textcolor{blue}{0.879}}  & \textbf{\textcolor{blue}{0.883}}  \\
		\hline
		\textbf{CLiF-VQA*}  & \textbf{Ours} &  \textbf{\textcolor{blue}{0.856}} & \textbf{\textcolor{blue}{0.860}}  & 0.894  & 0.895  & 0.868  & 0.870  \\
		\textbf{CLiF-VQA}  & \textbf{Ours} &  \textbf{\textcolor{red}{0.864}} & \textbf{\textcolor{red}{0.880}}  & \textbf{\textcolor{red}{0.902}}  & \textbf{\textcolor{red}{0.903}}  & \textbf{\textcolor{red}{0.888}}  & \textbf{\textcolor{red}{0.891}}  \\
		\toprule
	\end{tabular}
\end{table*}

\subsection{Pre-training Results on LSVQ}
We pre-train CLiF-VQA on the LSVQ dataset and compare it with the existing advanced classical and deep VQA methods on four test sets, as shown in Tab. \ref{table:1}.
All experiments are conducted under 10 train-test splits.
Compared with some classical methods (BRISQUE, TLVQM, VIDEVAL), CLiF-VQA achieves a significant improvement in performance on all test sets.
In addition, CLiF-VQA gives better results on all test sets compared to FAST-VQA, which focuses only on low-level-aware features of the video. This suggests that the introduced human perception features can well complement the features extracted by FAST-VQA, thus improving the prediction accuracy of the model.
Furthermore, CLiF-VQA has improved performance on the intra-dataset test sets compared to the current state-of-the-art DOVER (our reproduced). And CLiF-VQA also significantly outperforms the best results on LIVE-VQA in the cross-dataset test.

\subsection{Fine-tuning Results on Small Datasets}
In Tab. \ref{table:2}, we fine-tune CLiF-VQA on three small datasets (LIVE-VQC, KoNViD-1k, YouTube-UGC).
As before, all experiments are conducted under 10 train-test splits.
Here, we further test the results of CLiF-VQA fine-tuning on small data sets, when pre-training only the FAST-VQA branch on LSVQ.
It can be seen that CLiF-VQA* performs better than FAST-VQA on all three datasets.
It even surpasses DOVER on the LIVE-VQC dataset.
In addition, CLiF-VQA pretrained on LSVQ outperforms the current state-of-the-art DOVER (our reproduced) on all three datasets.

\begin{table}[h]
	\caption{Ablation study of three main components: Semantic (Semantic Feature Extraction Module), Spatial (Spatial Feature Extraction Module) and Regression (Regression Head). 
		SROCC and PLCC are the average results over the three datasets (LIVE-VQC, KoNViD-1k, YouTube-UGC).}
	\centering\setlength{\tabcolsep}{9.5pt}
	\label{table:3}
	\footnotesize
	\begin{tabular}{c cc|cc}
		\bottomrule
		Semantic & Spatial & Regression & SROCC & PLCC \\
		\hline
		\checkmark &   &   &  0.792 &  0.788 \\
		& \checkmark  &   &  0.864 &  0.865 \\
		\checkmark&  & \checkmark  &  0.812 &  0.820 \\
		& \checkmark & \checkmark  &  0.868 &  0.869\\
		\checkmark & \checkmark &   &  0.879 &  0.882\\
		\checkmark & \checkmark & \checkmark  &  0.885 &  0.889 \\
		\toprule
	\end{tabular}
\end{table}

\subsection{Ablation Studies}

\noindent
\textbf{Ablation on the Compositions of CLiF-VQA.} We validate the effectiveness of the three modules that make up CLiF-VQA. As shown in Tab. \ref{table:3}, CLiF-VQA has acceptable performance when only semantic features related to human feelings are extracted, and the performance of CLiF-VQA is further improved when the regression head module is introduced.
However, when only low-level perceptual features of the video are extracted using FAST-VQA, there is no significant improvement in performance after the introduction of the regression head.
When both branches are considered simultaneously, the performance of the model improves significantly relative to the use of a single branch and is further improved with the introduction of the regression header.

\begin{table}[h]
	\caption{Ablation study on descriptions. Obj and Sub denote objective descriptions and subjective descriptions, respectively.}
	\centering\setlength{\tabcolsep}{5pt}
	\label{table:4}
	\footnotesize
	\begin{tabular}{l|c|c|c}
		\bottomrule
		Datasets & LIVE-VQC & KoNViD-1k & YouTube-UGC  \\
		\hline
		Descriptions & SROCC/PLCC & SROCC/PLCC & SROCC/PLCC  \\
		\hline
		None & 0.845/0.852 & 0.890/0.889 & 0.857/0.853 \\
		Only-Obj & 0.857/0.868 & 0.898/0.895 & 0.880/0.876 \\
		Only-Sub & 0.849/0.856 & 0.893/0.891 & 0.863/0.860  \\
		Obj+Sub & 0.864/0.873 & 0.902/0.903 & 0.888/0.891  \\
		
		\toprule
	\end{tabular}
\end{table}

\noindent
\textbf{Ablation on Descriptions.} In Tab. \ref{table:4}, we verify the effect of different types of descriptions on the performance of CLiF-VQA.
The results demonstrate that objective descriptions have a greater impact on the performance improvement of CliF-VQA compared to subjective descriptions.
And the optimal results can be obtained by using both objective and subjective descriptions.

\noindent
\textbf{Ablation on the Number of Samples.} We use CLiF-VQA pre-trained on LSVQ for cross-dataset testing.
We divide LIVE-VQC into three groups according to the resolution of the videos: 1080p (110 videos), 720p (316 videos), and 540p (159 videos). We then evaluated the performance on these three resolution groups with different numbers of samples (Fig. \ref{fig6-b}).  As shown in Tab. \ref{table:6}, CLiF-VQA achieves optimal performance with a smaller number of feature extractions on low-resolution videos, while more feature extractions are required for high-resolution videos.
It also shows that the 9 times feature extraction performed in this paper does not show the optimal performance of CLiF-VQA.
Therefore, the performance of CLiF-VQA proposed in this paper can be further improved by increasing the number of samples.

\begin{table}[t]
	\caption{Experimental performance regarding the number of samples taken by the semantic feature extractor on video frames of different resolutions.
	Videos in all resolutions are from LIVE-VQC.}
	\centering\setlength{\tabcolsep}{6.5pt}
	\label{table:6}
	\footnotesize
	\begin{tabular}{c|c|c|c}
			\bottomrule
			Resolution & $ \le $540p & 720p & 1080p \\
			\hline
			Times & SROCC/PLCC & SROCC/PLCC & SROCC/PLCC  \\
			\hline
			1 & 0.841/0.876 & 0.811/0.823 & 0.812/0.809   \\
			5 & 0.856/0.884 & 0.825/0.840 & 0.820/0.822   \\
			9 & 0.862/0.889 & 0.843/0.858 & 0.834/0.836  \\
			25 & \textbf{0.866/0.890} & 0.852/0.866 & 0.847/0.858   \\
			45 & 0.864/0.889 & \textbf{0.855/0.868} & 0.855/0.867   \\
			85 & 0.864/0.887 & 0.851/0.862 & \textbf{0.857/0.871}   \\
			\toprule
		\end{tabular}
\end{table}

%% file: sec/6_conclusions.tex
\section{Conclusions}
\label{sec:6}
In this paper, we first analyze that human subjective feelings have an important impact on video quality evaluation.
Further, we verify for the first time the correlation between CLIP and human feelings in video perception.
Extensive subjective experiments demonstrate that CLIP not only has good consistency with human feelings, but also can achieve satisfactory results in VQA by using only the features related to human feelings extracted by CLIP.
Motivated by these findings, we propose CLiF-VQA, a method that extracts features related to human feelings and low-level perceptual features of videos. Then, it fuses the two features to obtain the video quality score. Extensive experimental results demonstrate that the proposed CLiF-VQA outperforms existing methods on multiple VQA datasets.